%% file: ms.tex
\documentclass{article}

\usepackage[final,nonatbib]{ms}

\usepackage[utf8]{inputenc} 
\usepackage[T1]{fontenc}    
\usepackage{hyperref}       
\usepackage{url}            
\usepackage{booktabs}       
\usepackage{amsfonts}       
\usepackage{nicefrac}       
\usepackage{microtype}      

\usepackage{graphicx}
\usepackage{subcaption}
\input{math_commands.tex}

\usepackage{gensymb}
\usepackage{float}

\usepackage[hyphenbreaks]{breakurl}

\title{Predicting the Solar Potential of Rooftops using Image Segmentation and Structured Data}

\author{%
Daniel DE BARROS SOARES \\
namR, Paris, France\\
daniels@namr.com

\And
François ANDRIEUX \\
namR, Paris, France\\
francoisa@namr.com
\And
Bastien HELL \\
namR, Paris, France\\
bastienh@namr.com
\And
Julien LENHARDT \\
ENSTA Paris \\
namR, Paris, France\\
julien.lenhardt@ensta-paris.fr
\And
Jordi BADOSA \\
LMD, Ecole polytechnique, IP Paris \\
Palaiseau, France \\  
jordi.badosa@lmd.polytechnique.fr
\And
Sylvain GAVOILLE \\
namR, Paris, France \\
sylvaing@namr.com
\And
Stéphane GAIFFAS \\
LPSM, Université de Paris\\ 
DMA, Ecole normale supérieure \\
namR, Paris, France\\
stephaneg@namr.com
\And
Emmanuel BACRY \\\
CEREMADE, Université Paris Dauphine\\ 
namR, Paris, France\\
emmanuelb@namr.com
  
}

\begin{document}

\maketitle
\begin{abstract}
 Estimating the amount of electricity that can be produced by rooftop photovoltaic systems is a time-consuming process that requires on-site measurements, a difficult task to achieve on a large scale. In this paper, we present an approach to estimate the solar potential of rooftops based on their location and architectural characteristics, as well as the amount of solar radiation they receive annually. Our technique uses computer vision to achieve semantic segmentation of roof sections and roof objects on the one hand, and a machine learning model based on structured building features to predict roof pitch on the other hand. We then compute the azimuth and maximum number of solar panels that can be installed on a rooftop with geometric approaches. Finally, we compute precise shading masks and combine them with solar irradiation data that enables us to estimate the yearly solar potential of a rooftop.

\end{abstract}

\section{Introduction}
\label{sec:introduction}

The 21st century is characterized by an ever-increasing energy consumption and greenhouse gas emissions that are contributing to climate change in an unprecedented way. Fossil fuels are still our main source of electricity and heat generation, accounting for 42\% of the greenhouse gas emissions in 2016 \cite{iea_co2_emission_2018}. Energy efficiency and development of renewable energies are presented as the two main approaches to lower these emissions \cite{irena_2019} and photovoltaic solar energy is one of the fastest-growing renewable energy sources because of low maintenance and operation costs \cite{irena_renewable_cost_2017}. Solar panels can also produce energy anywhere there is enough sunlight without direct impact on the  environment, giving the opportunity to produce energy in dense environments such as cities or industrial zones. 

\begin{figure}
\begin{center}
\includegraphics[width=\textwidth]{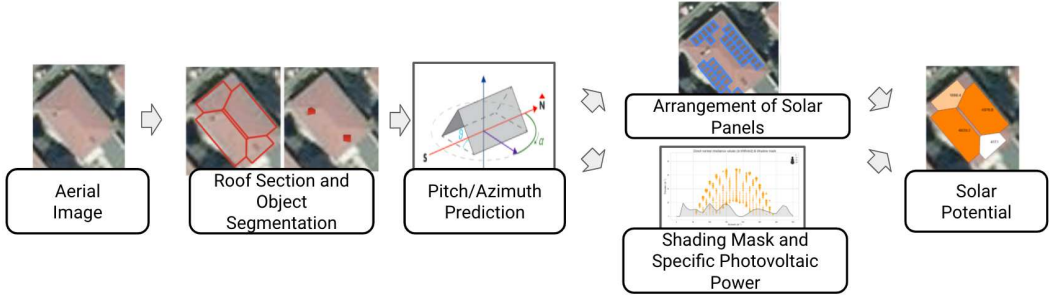}
\caption{\label{fig:fig_1} Workflow for a given building from the initial aerial image to its rooftop solar potential.}
\end{center}
\end{figure}

In order to compute the solar potential of a building rooftop, we need two types of information: how many solar modules can be fitted on a roof section and how much energy each of this module could produce within a year, accounting for the local irradiation and shading. The present work proposes a methodology to answer these questions, as illustrated in \textbf{Figure \ref{fig:fig_1}}. 

First, an image segmentation algorithm applied to aerial imagery is used to obtain each roof section 2D geometry and any obstructing object it may contain (\textbf{Section \ref{subsec:segmentation}}). The roof 3D geometry (pitch/azimuth) is obtained through a combination of geometric methods and a Random Forest algorithm (\textbf{Section \ref{subsec:pitch_azimuth}}). Then, the maximum number of modules fitting on a section is computed using a geometric packing algorithm (\textbf{\hyperref[sec:annex]{Appendix}}). A shading mask is computed from the shadows casted by surrounding buildings and relief and the specific photovoltaic power is computed from meteorological data, module orientation and shading effects  (\textbf{Section \ref{subsec:shading_pvout}}). Finally, the total amount of energy produced by a given roof section in a year can be computed by multiplying the number of modules $N_{modules}$, the module's nominal power $P_{max}$ and its specific photovoltaic power
$PV_{out}$:

$$
solar \: potential \: (kWh/year) = N_{modules}*P_{max}*PV_{out}
$$

\section{Methodology}

\subsection{Roof and object segmentation} \label{subsec:segmentation}

\begin{figure}[H]
\begin{center}
\includegraphics[width=\textwidth]{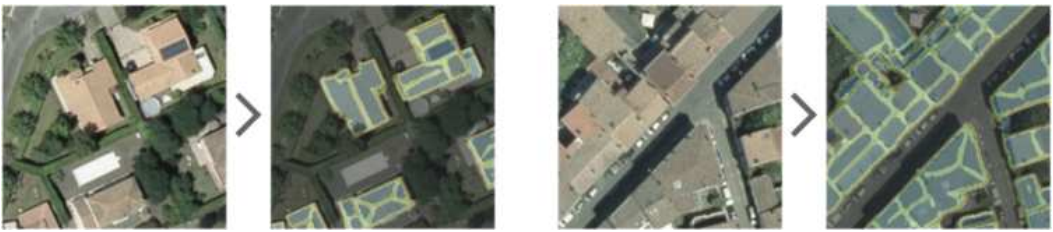}
\caption{Two examples of aerial image and roof sections semantic segmentation pairs. Sections are depicted in blue, ridges in yellow.}
\label{fig:fig_2}
\end{center}
\end{figure}

The first step in our pipeline is to subdivide roof sections as well as the equipments present on it. This will help us understand how much space is available on a roof for it to be equipped with solar panels. We split this task into two semantic segmentation tasks, consisting in classifying each pixel of an image into one of multiple classes. Here, we developed one model that segments the images into \textit{background}, roof \textit{sections} and roof \textit{ridges} as depicted in \textbf{Figure \ref{fig:fig_2}}, and one that subdivides the images into \textit{background} and a set of chosen roof objects. Roof sections are eventually enhanced by geometrical regularization (see \textbf{\hyperref[sec:annex]{Appendix}}).

Both segmentation models consist in a U-Net \cite{unet} architecture using a ResNet-34 \cite{resnet} feature extractor as backbone. The main dataset used in this work is the set of aerial images obtained from the IGN (French National Institute of Geographic and Forest Information) dataset \cite{bdortho}. Building footprints were also obtained from the same source \cite{bdtopo}. Roof geometries labels are obtained from the 3D city representations of five different French cities. No roof object dataset was found and we relied instead on a set of 423 manually tagged images including object type (\textit{smoke vent}, \textit{roof window}, etc.) and geometry. 

\subsection{Pitch and azimuth prediction} \label{subsec:pitch_azimuth}

Transforming the 2D prediction of a roof to its 3D representation requires estimating its pitch and azimuth. The pitch represents the roof inclination ranging from $0^{\circ}$ (flat roof) to $90^{\circ}$ (a theoretical vertical roof). We predict a roof's mean pitch as a linear function of the latitude and use a machine learning method to predict the normalized pitch, defined as $pitch_{norm}=(pitch - pitch_{mean})/pitch_{mean}$. Training data comes from the same five French cities as in \textbf{Section \ref{subsec:segmentation}} and a Random Forest with 100 trees and a max depth of 15 was used for the normalized pitch using features such as roof material, roof type, building height and roof shape.

The azimuth corresponds to the roof's orientation and is computed using a purely geometric algorithm. First, we compute $\theta_{bb}$, the orientation of the roof's bounding box modulo $90^{\circ}$. We then make the hypothesis that the roof azimuth has the form   $\theta_{bb} + \Delta\theta$, where $\Delta\theta$ can have one of the following 4 values : $0^{\circ}$, $90^{\circ}$, $180^{\circ}$ or $270^{\circ}$. This gives us a 4 classes classification problem where we want to predict $\Delta\theta$. Finally, we use nearby roof sections as an indicator of the correct orientation, under the hypothesis that roof sections tend to be  oriented away from nearby roofs. This can be observed for the hipped roof in \textbf{Figure \ref{fig:fig_1}} where each section has a different orientation, facing away from each other.

\subsection{Specific photovoltaic power and shading mask} \label{subsec:shading_pvout}

Lastly, the specific photovoltaic power output $PV_{out}$ is calculated as the average of the yearly production normalized to 1 kWp of installed capacity. Multi-year hourly irradiance estimations and meteorological variables are used as input on a latitude/longitude spatial grid. Analytical equations from the pvlib python library \cite{william_pvlib} are then used to transpose the direct and diffuse irradiance components onto the plane of the rooftop and to estimate the PV panel temperature. These two variables are then used to compute the $PV_{DC}$ power using the PVWatts model \cite{pvwatts}. The AC power is computed by using considered inverter specifications and system losses are applied. 

Since the performance of photovoltaic installations is deeply impacted by complete or partial shading of the photovoltaic cells \cite{dolara2013experimental} \cite{alonso2005solar} \cite{silvestre2007effects}, it is important to take shading into account when computing the roof's solar potential. Several approaches to factor shading and assess the solar potential of buildings have already been described \cite{freitas2015modelling} \cite{desthieux2018solar}. We herein compute a shading mask for the whole roof based on the Sky-View factor methodology \cite{zaksek2011sky} and consider two different scales in order to account for shading from neighboring buildings and the surrounding topography. More details on this method are given in \textbf{\hyperref[sec:annex]{Appendix}}.

Finally, $PV_{out}$ is computed as the annual integral of the obtained AC power hourly values. Before this integration, the  hourly power contribution from direct and diffuse solar irradiance are given separately so that shading can be applied to these two components. 


\section{Results}
\label{sec:results}

In this section we describe the pipeline's results for each individual part. As we could not find an open dataset for our scope (French territory) that would allow us to validate the entire workflow, we adopted the strategy of validating separately the algorithms used in each step, as described below.

The roof segmentation reached a pixel-wise accuracy of 77\% on roof section segmentation (excluding the background), and 30\% on roof objects. The large gap in performance is explained by the difference in training data. As opposed to the small set of manually labeled roof objects, many accurate roof sections were extracted from the cities' 3D models. This led to large amounts of accurate roof sections labels as well as estimations for pitch and azimuth. Despite being identified by the model, roof objects are poorly segmented and the confusion between classes is high. The detection and segmentation of roof objects is a task we are working on actively, labeling more data with a broader range of objects types.

\begin{table}[H]
\begin{center}
\begin{tabular}{|c|c|c|}
  \hline
 Task & Model & Score \\
  \hline
  Roof sections segmentation & ResNet-34-based UNet & Pixel accuracy = 77\% \\
  Roof objects segmentation & ResNet-34-based UNet & Pixel accuracy = 30\% \\
  Azimuth & Geometric & Accuracy = 79\% \\
  Mean pitch as a function of latitude & Linear Regression & ${R}^2$ = 0.93, MAE = $3.9^{\circ}$ \\
  Normalized pitch  & Random Forest & ${R}^2$ = 0.37, MAE = $5.5^{\circ}$ \\
  \hline
\end{tabular}
\end{center}
\caption{Our methodology scores for algorithms where a validation set is available. Two different tasks are used to predict the pitch.}
\label{tab:pitch_azimuth}
\end{table}

\begin{figure}[ht]
\centering
\begin{subfigure}[b]{.45\textwidth}
\centering
\includegraphics[height=4.5cm]{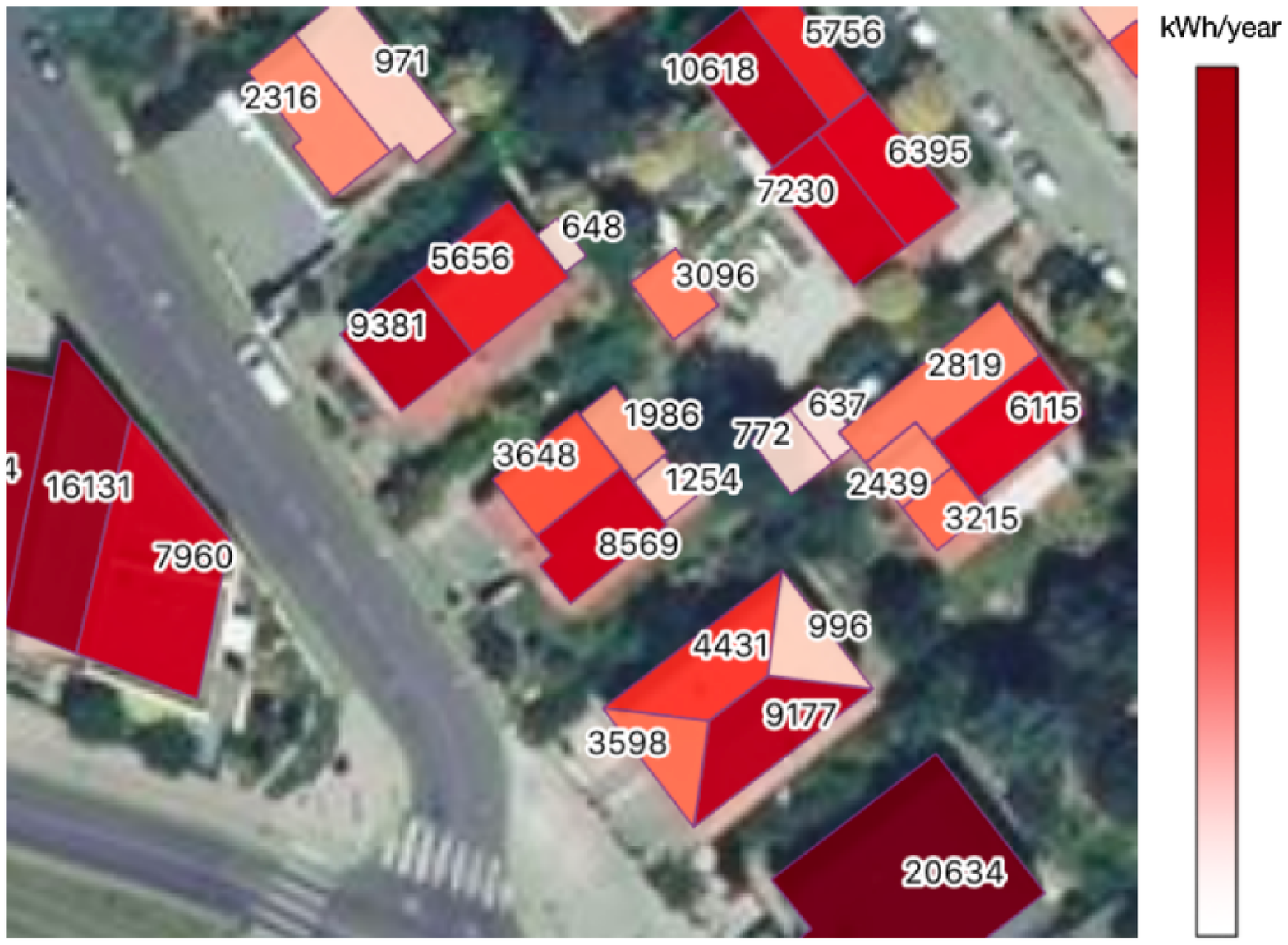}
\caption{}
\label{fig:fig_3a}
\end{subfigure}
~
\begin{subfigure}[b]{.45\textwidth}
\centering
\includegraphics[height=4.5cm]{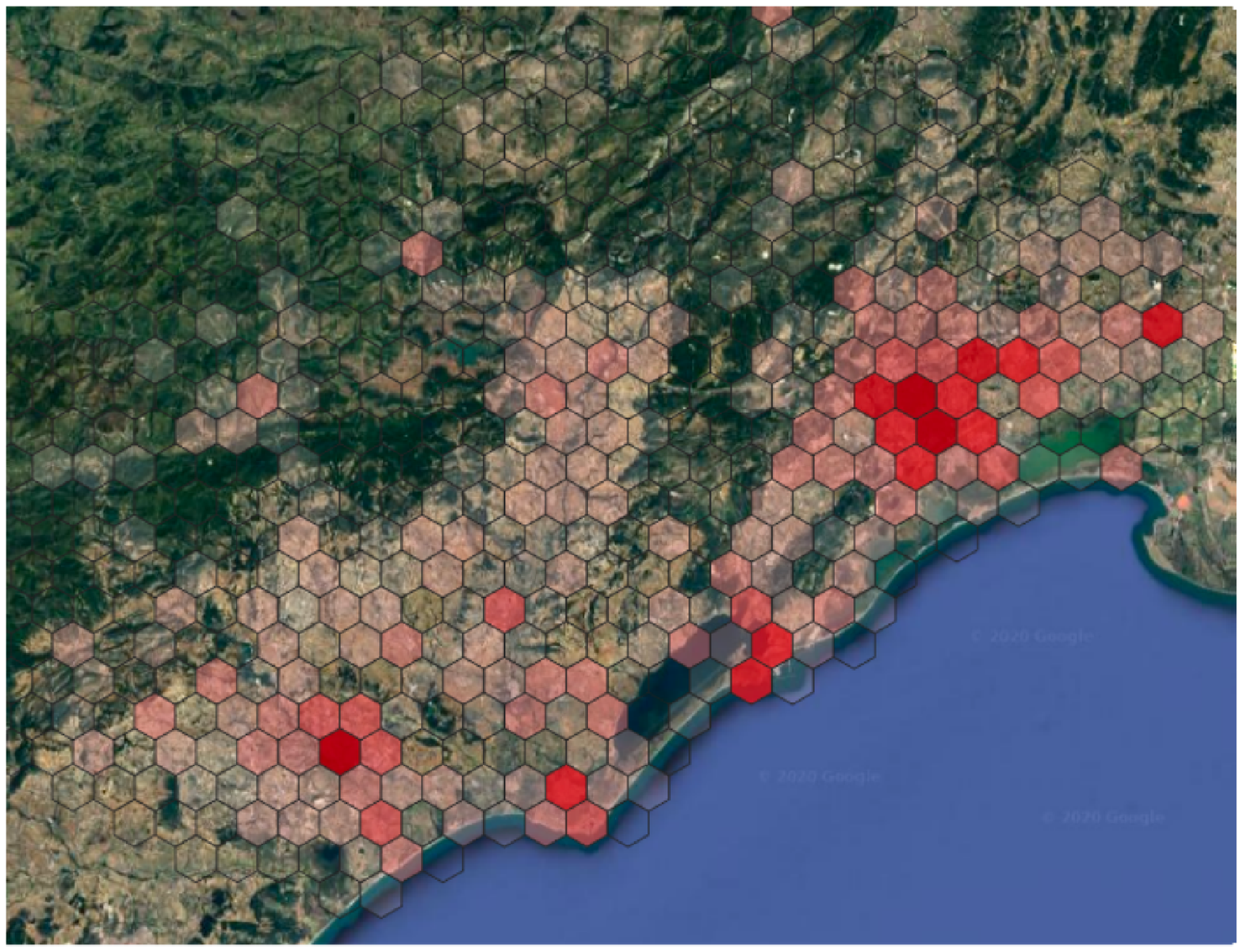}
\caption{}
\label{fig:fig_3b}
\end{subfigure}
\caption{(a) Neighborhood view of the resulting solar potential. (b) Aggregated view (sum) of the solar potential over French Hérault department. }
\label{fig:fig_3}
\end{figure}

The score for the roof section pitch linear regression task was obtained by fitting the mean values of four cities and predicting the fifth one. The final score is obtained by averaging all the five combinations and presents a rather high ${R}^2$ value of 0.93. This indicates a good correlation but the small amount of latitude values implies that we should be careful when extrapolating these results. The normalized pitch model has an MAE of 0.2 corresponding to an absolute pitch MAE of $5.5^{\circ}$ (averaging across the five cities). The azimuth model reached good results with an accuracy of 79\%. We can put into perspective the impacts of pitch and azimuth values on the final solar potential of a rooftop. Using a rooftop located in Montpellier ($43^{\circ}36'$, $03^{\circ}52'$), South-facing and with an optimal pitch ($37^{\circ}$) as a reference, we can compute the impacts of pitch/azimuth variation using \cite{globalsolaratlas}. We see that a $10^{\circ}$ pitch variation reduces the solar potential by 1\% whereas using an East-facing rooftop reduces it by 25\%. \textbf{Figure \ref{fig:fig_3a}} illustrates this variation.

No open-access validation dataset was found for the shading impact on the solar potential. We adopted instead a qualitative validation comparing urban areas with different building densities and regions with different topographic surroundings (data not shown). The same qualitative approach was adopted for the resulting solar potential as we also lack large-scale  validation data.

Eventually, we estimated the solar potential of 3.5 millions roof sections accounting for more than 1.1 million buildings. \textbf{Figure \ref{fig:fig_3b}}. illustrates the solar potential over French Hérault department.

\section{Conclusion}

In this paper, we described a complete functional pipeline used to predict the yearly solar potential of rooftops using aerial imagery, building features and open data labels. The main source of ground truth comes from the open-access building 3D geometries of five French cities. The methodology entirely relies on structured data and aerial imagery that is available at scale, which will ultimately enable us to predict France’s solar potential on every roof.

The main limitation of the methodology described in this paper is the lack of validation datasets for the entire pipeline and, in particular, for the shading algorithm. We adopted the strategy of validating separately the algorithms used in each step, as described in \textbf{Section \ref{sec:results}}. The methodology developed here is intended for a commercial application and, until now, our main validation came from our clients in the energy sector that compare our results to their internal data. 

The two main steps of our methodology, roof section segmentation and azimuth prediction, show very good results and are well-adapted to our present application. Roof object segmentation and pitch prediction however, present rather poor results. The first one is caused by a lack of high-quality labeled data. On the other hand, the poor results of the pitch prediction step are mainly caused by the fact that the pitch cannot be easily estimated from aerial imagery and has little relationship with building features. As discussed in \textbf{Section \ref{sec:results}}, this limitation is compensated by the low impact the pitch has on the final solar potential compared to other features such as the azimuth and the roof section surface.

We hope that our methodology will contribute to a better understanding of the energy potential achievable with the massive installation of solar panels on residential and commercial buildings in order to accelerate the sustainable energy transition.

\section*{Appendix}
\label{sec:annex}

\subsection*{Existing commercial solutions}
Some commercial solutions that estimate rooftop solar potential data for a given building already exist. The North American technology company Google, through its project Sunroof, proposes a cadastre in a large part of the United States territory \cite{cadastre_sunroof} but also in main French cities in collaboration with the French energy group ENGIE \cite{cadastre_engie}. The French start-up In Sun We Trust, in collaboration with the Swedish energy company Otovo, covers the entire French territory \cite{cadastre_insunwetrust}. Others French companies also have a cadastre for a smaller scope, such as Rhino Solar \cite{cadastre_rhino} in the Lyon region and Cythelia Energy \cite{cadastre_cythelia} for on-demand scopes. Each one of these solutions has different levels of complexity on its methodology and results. However, their data and methods cannot be directly accessed and we cannot compare them to the methodology we showcase here.

\subsection*{Roof section regularization}
\label{subsec:regularization}

\begin{figure}[H]
\includegraphics[width=\textwidth]{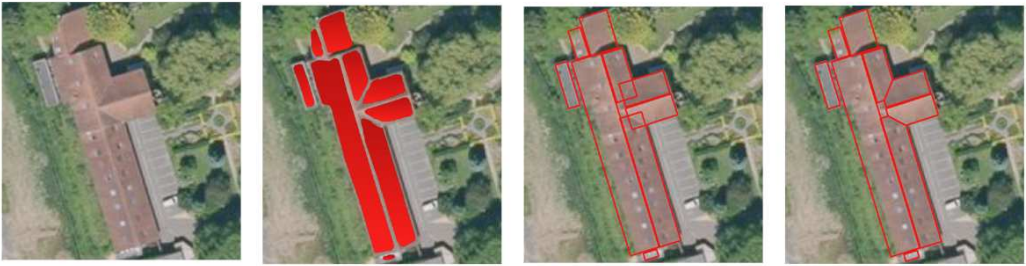}
\caption{Illustration of the roof vectorization and regularization process. From left to right: raw aerial image, U-Net segmentation, oriented bounding boxes and fitted predictions using geometric approaches.}
\label{fig:fig_4}
\end{figure}

Roof and objects segmentation models output pixel-wise predictions, which are then vectorized and stored into our database. Roof sections polygons undergo further post-processing and are regularized in order to fit the buildings' footprints. We compute the oriented bounding box of each roof section according to its supporting building's facade, and cut overlapping boxes along their intersection axis. This process is shown in \textbf{Figure \ref{fig:fig_4}}.

\subsection*{Arrangement of solar panels} \label{subsec:packing}

\begin{figure}[H]
\begin{center}
\includegraphics[width=\textwidth]{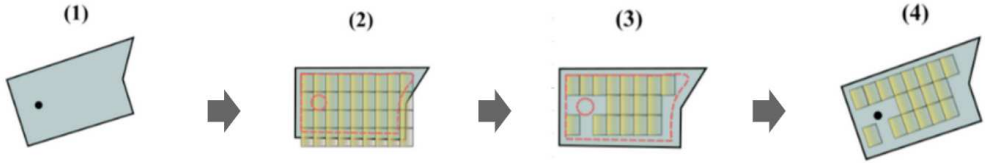}
\caption{Illustration of the greedy packing algorithm. (1) depicts a roof section containing an object. (2) is the panel packing within the eroded shape after rotating the roof section. (3) shows the best configuration taking into account the margins. (4) depicts the result of the algorithm.}
\label{fig:fig_5}
\end{center}
\end{figure}

Given the roof shape, the presence of objects and the dimensions of the chosen solar module, we are able to compute the maximum number of modules per roof section. We developed a greedy geometric algorithm that tries to fit the maximum number of panels per row before dropping those that intersect the roof's ridges, boundaries or eventual obstructing object. The minimum distance between each panel and the roof borders is also taken into account, as shown in \textbf{Figure \ref{fig:fig_5}}.

\subsection*{Shading} 
\label{subsec:shading}

\begin{figure}[ht]
\centering
\begin{subfigure}[b]{.2\textwidth}
\centering
\includegraphics[width=\textwidth]{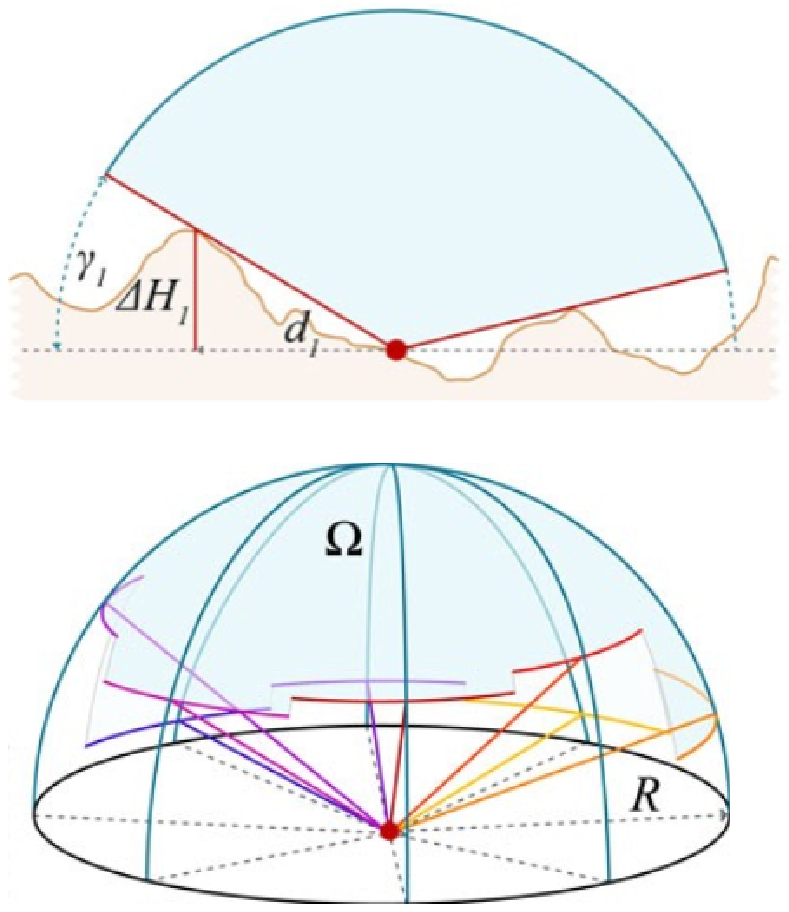}
\caption{}
\label{fig:fig_6a}
\end{subfigure}
~
\begin{subfigure}[b]{.3\textwidth}
\centering
\includegraphics[width=\textwidth]{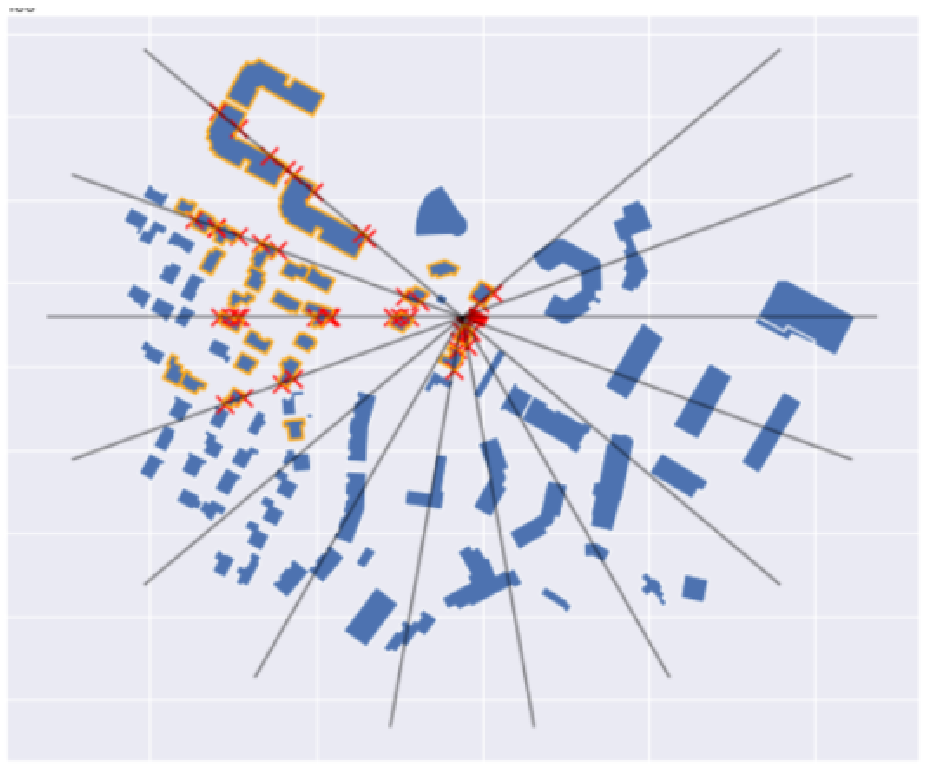}
\caption{}
\label{fig:fig_6b}
\end{subfigure}
~
\begin{subfigure}[b]{.45\textwidth}
\includegraphics[width=\textwidth]{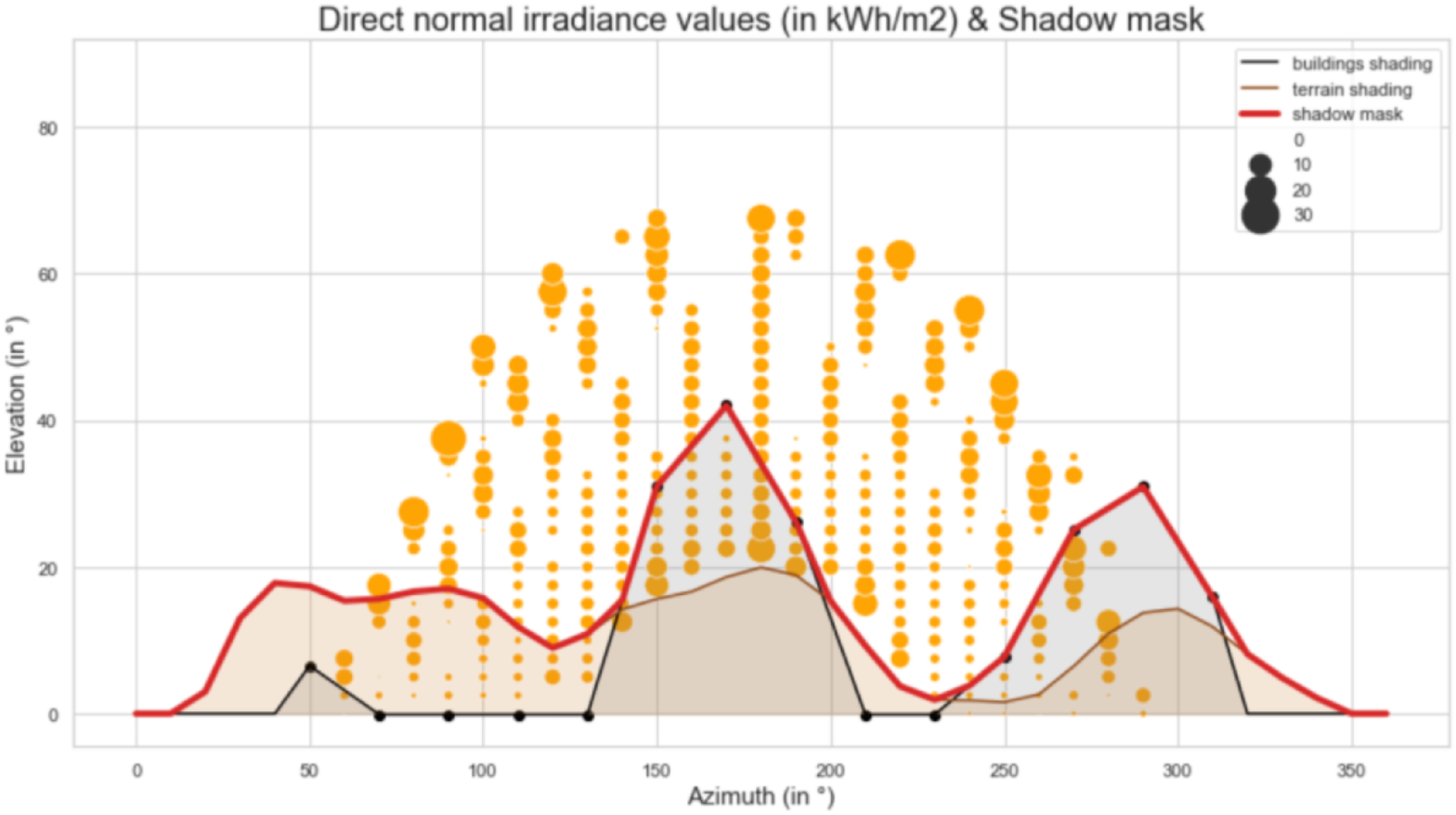}
\caption{}
\label{fig:fig_6c}
\end{subfigure}
\caption{(a) Sky-View-Factor as the area above the obstacles $\Omega$ over the total area of the semi-sphere; (b) Rays intersecting the surrounding buildings footprints in each of the sampled azimuth directions, pre-filtered buildings are outlined in orange; (c) Shading masks combined and yearly direct normal irradiance map.}
\label{fig:fig_6}
\end{figure}

The first part of the shading mask computation is performed based on the buildings' geometry and height by dividing the azimuth range in regular sections, ray tracing and marking the intersections with the buildings' footprint, as developed in \cite{dorman2019shadow} and illustrated in \textbf{Figure \ref{fig:fig_6b}}. To speed up computations, we can choose a smaller angular resolution and apply a distance and height-based pre-filtering of the buildings we consider. The second part uses a Digital Elevation Model from the IGN \cite{bdalti} and projected shadows computations in QGIS \cite{qgis}, \cite{qgis_landcaspe} to produce a similar shading mask for each cell of the input raster. This method is similar to the ones developed from image processing in \cite{richens1997image}, \cite{richens1999urban} and \cite{stewart1998fast} for casting shadows.

By combining shading masks as described above, we are ultimately able to model the presence of shading on the rooftop and consequently adapt the irradiation values used to compute the solar potential, as shown in \textbf{Figure \ref{fig:fig_6c}}.

\subsection*{Specific photovoltaic power} \label{subsec:pvout}

\begin{figure}[H]
\includegraphics[width=\textwidth]{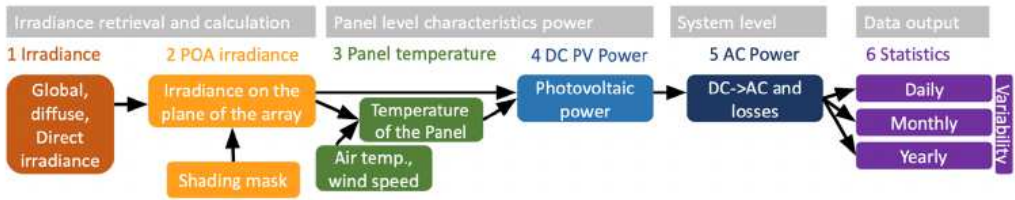}
\caption{Step-by-step methodology to compute $PV_{out}$.}
\label{fig:fig_7}
\end{figure}

The steps used to compute the specific photovoltaic power, as described in \textbf{Section \ref{subsec:shading_pvout}}, is shown in \textbf{Figure \ref{fig:fig_7}}

\end{document}

%% file: math_commands.tex
\usepackage{amsmath,amsfonts,bm}

\def\eqref#1{equation~\ref{#1}}

\def\1{\bm{1}}

\DeclareMathAlphabet{\mathsfit}{\encodingdefault}{\sfdefault}{m}{sl}
\SetMathAlphabet{\mathsfit}{bold}{\encodingdefault}{\sfdefault}{bx}{n}

